\documentclass[letterpaper]{article} 
\usepackage[]{aaai2026}  
\usepackage{times}  
\usepackage{helvet}  
\usepackage{courier}  
\usepackage[hyphens]{url}  
\usepackage{graphicx} 
\usepackage{natbib}  
\usepackage{caption} 
\usepackage{algorithm}
\usepackage{amsmath}
\usepackage{amssymb}
\usepackage{bm} 
\usepackage{algorithmicx}
\usepackage{algpseudocode}
\usepackage{booktabs}
\usepackage{multirow}

\usepackage{newfloat}
\usepackage{listings}
\DeclareCaptionStyle{ruled}{labelfont=normalfont,labelsep=colon,strut=off} 
\floatstyle{ruled}
\newfloat{listing}{tb}{lst}{}
\floatname{listing}{Listing}
\title{Self-similarity Analysis in Deep Neural Networks}

\author {
    Jingyi Ding\textsuperscript{1}\textsuperscript{*}, 
  Chengwen Qi\textsuperscript{1}, 
  Hongfei Wang\textsuperscript{1}, 
  Jianshe Wu\textsuperscript{1}, 
  Licheng Jiao\textsuperscript{1}, 
  Yuwei Guo\textsuperscript{1},
  Jian Gao\textsuperscript{1}
}
\affiliations {
    \textsuperscript{\rm 1}School of Artificial Inteligence, Xidian University,China\\

    jyding@xidian.edu.cn,24171214007@stu.xidian.edu.cn
}

\begin{document}

\maketitle

\begin{abstract}
Current research has found that some deep neural networks exhibit strong hierarchical self-similarity in feature representation or parameter distribution. However, aside from preliminary studies on how the power-law distribution of weights across different training stages affects model performance, there has been no quantitative analysis on how the self-similarity of hidden space geometry influences model weight optimization, nor is there a clear understanding of the dynamic behavior of internal neurons. Therefore, this paper proposes a complex network modeling method based on the output features of hidden-layer neurons to investigate the self-similarity of feature networks constructed at different hidden layers, and analyzes how adjusting the degree of self-similarity in feature networks can enhance the classification performance of deep neural networks. Validated on three types of networks—MLP architectures, convolutional networks, and attention architectures—this study reveals that the degree of self-similarity exhibited by feature networks varies across different model architectures. Furthermore, embedding constraints on the self-similarity of feature networks during the training process can improve the performance of self-similar deep neural networks (MLP architectures and attention architectures) by up to 6\%.
\end{abstract}


\section{Introduction}
Neural network models, especially deep neural networks, have achieved remarkable success across various domains due to their powerful representation learning capabilities \cite{lecun2015deep}. Numerous studies currently leverage complex network theory \cite{la2024deep} to explain the dynamic behavior of internal neurons, with particular emphasis on understanding structural properties such as small-worldness \cite{watts1998collective}, scale-freeness \cite{barabasi1998emergence}, and self-similarity \cite{song2005self}. These topological characteristics hold significant importance for enhancing model interpretability, improving performance, and designing novel architectures.

Traditional approaches typically directly map a model's connection onto a complex network and subsequently analyze its topological properties \cite{testolin2020deep}\cite{zambra2020emergence}. However, this methodology overlooks the dynamic structural changes induced by model training and fails to capture the deep information associations embedded in neuronal activation states. Furthermore, given the substantial architectural differences among neural network models, the resulting complex network connection patterns vary considerably. This diversity highlights the lack of a universal, standardized method for analyzing and comparing the intrinsic network properties across different structures.

Therefore, this paper proposes a new perspective for studying the internal mechanisms of deep network models. By utilizing the output features of hidden-layer neurons during the feedforward process, we construct a dynamic complex network that reflects the intrinsic information relationships within the model.

Self-similarity, as one of the core topological characteristics of complex networks, originates from the intersection of fractal geometry and statistical physics. It manifests as scale invariance in the statistical properties between network substructures and the whole network \cite{song2005self}. In the field of neural networks, research on self-similarity reveals two primary trends: structural embedding and data-driven approaches. Some studies attempt to deliberately integrate self-similarity into network architectures through fractal stacking \cite{larsson2016fractalnet}, hierarchical connection control \cite{dorogov2021morphological}\cite{dorogov2021stratified}, or modular repetition \cite{krizhevsky2012imagenet} to construct models with inherent self-similar properties. These methods demonstrate that structural self-similarity may benefit model performance.

Other research primarily explores leveraging the self-similarity inherent in data itself to facilitate model learning \cite{wewer2023simnp}\cite{liang2020video}. However, both structural embedding and data-driven approaches lack in-depth analysis of how the intrinsic self-similarity of neural networks dynamically influences the optimization process of parameters. Therefore, this paper designs a self-similarity evaluation metric SS\_rate for the $G_M$ of hidden-layer feature networks. We observe its variations during training and incorporate it into the loss function as a differentiable regularization term to dynamically constrain and guide the model learning process.
In summary, the contributions of our work are
\begin{itemize}
    \item Proposes an architecture-agnostic complex network modeling method based on hidden-layer features that captures parameter dynamics.
    \item Designs SS\_rate — a differentiable and computationally efficient self-similarity metric for feature networks ($G_M$) — enabling its direct integration into loss functions to dynamically constrain and optimize model self-similarity.
    \item Reveals distinct $G_M$ visualization patterns across diverse deep neural network models, linking these variations to their computational paradigms; 
    \item Observes a pervasive degradation of self-similarity in feature networks during training; and demonstrates that SS\_rate-based regularization enhances performance in networks with intrinsic self-similar characteristics.
\end{itemize}

\section{Related Work}

\textbf{Complex Network Properties of Neural Networks}: Biological studies \cite{monteiro2016model}  identify small-worldness and scale-freeness as critical topological properties governing learning efficiency in biological neural networks, revealing their essential role in accelerating knowledge acquisition.\cite{erkaymaz2017performance} Research on feedforward networks (FFNs) demonstrates that small-world architectures actively enhance model performance by optimizing information propagation pathways.\cite{you2020graph} Adopting an approach similar to ours, this work models neural networks as complex systems and identifies optimal topological configurations through systematic screening of clustering coefficients and average path lengths.\cite{xie2019exploring} By generating neural architectures via parametric complex network generators, this study establishes quantifiable correlations between generator parameters, topological features, and model performance.\cite{scabini2023structure}A graph-based representation for fully connected networks is proposed, employing complex network theory to analyze structure-function relationships. This approach reveals emergent topological patterns under diverse weight initialization schemes.\cite{la2021characterizing} This work introduces specialized metrics (e.g., node strength, layer-wise fluctuations) for analyzing fully connected and convolutional networks. These tools provide critical insights into structural evolution during training processes.

\noindent
\textbf{Self-Similarity in Neural Networks}:Current research predominantly focuses on utilizing the intrinsic connection structures of neural networks as the basis for investigating network properties. Key advancements in this domain include:\cite{larsson2016fractalnet}introduces fractal blocks that recursively combine shallow sub-networks to construct ultra-deep architectures, eliminating the need for residual connections.\cite{dorogov2021morphological} and \cite{dorogov2021stratified} jointly establish a theoretical-design-application framework for self-similar neural networks. Inspired by biological modularity, these works transform morphological principles into computational engineering solutions through stratified modeling, providing new paradigms for efficient training and hardware implementation.\cite{han2021transformer} decomposes images into "visual sentences" (image patches) and "visual vocabulary" (sub-patches), employing dual-layer Transformers to simultaneously model local details and global relationships. This fractal-inspired approach creates intrinsically self-similar architectures.
A distinct research direction investigates self-similarity within training data rather than network structures:\cite{wewer2023simnp} pioneers a learnable self-similarity prior for neural representations, establishing a novel methodological foundation for 3D reconstruction.\cite{liang2020video} develops a video super-resolution reconstruction method leveraging spatio-temporal feature self-similarity, simultaneously enhancing video quality while optimizing temporal efficiency and eliminating frame-jitter artifacts.\cite{lee2023revisiting} proposes a unified approach that captures internal image structures while progressively compressing them into dense self-similarity descriptors.

\section{Method}
\begin{figure*}[t]
\centering
\includegraphics[width=1\textwidth]{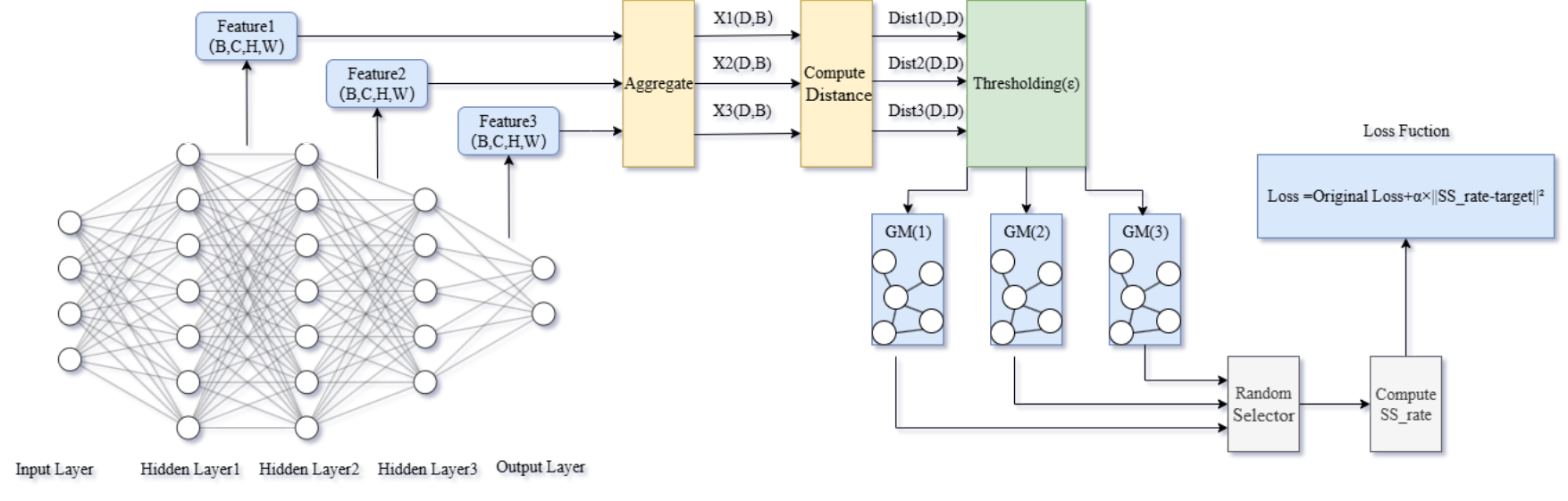} 
\caption{Application workflow of the proposed method on a three-layer fully connected neural network. The diagram illustrates feature extraction, distance computation, $G_M$ processing, and loss integration mechanisms.}
\label{fig:model}
\end{figure*}

To investigate the internal working mechanisms of neural network models, this paper proposes a dynamic complex network modeling method based on the output features of hidden-layer neurons. This approach studies the self-similarity of feature networks constructed in different hidden layers, starting from the perspective of how intrinsic self-similarity dynamically influences the optimization process of network parameter learning. We design a self-similarity metric SS\_rate as a differentiable regularization term to constrain model training, thereby controlling the model optimization process. The specific framework is shown in Figure~\ref{fig:model}.

\subsection{Building Feature Networks $\mathbf{G_M}$}
$G_M$ is a graph structure constructed from the feedforward hidden-layer features of neural network model $M$, which encodes both the inherent data patterns of the model's latent representations and intrinsically embeds its internal data processing mechanisms. The implementation workflow is detailed in Algorithm~\ref{alg:topology_graph}.

\begin{algorithm}[htb]
\caption{Topology Graph $G_M$ Construction}
\label{alg:topology_graph}
\begin{algorithmic}[1]
\Require Neural network model $M$, input data $X \in \mathbb{R}^{B}$
\Ensure Feature matrix $F \in \mathbb{R}^{(D,B)}$, topology graph $G_M$

\State \textrm{Compute feedforward hidden features}  
    \Statex \quad $H_1(B, d_0, d_1, \ldots, d_n) \leftarrow M(X)$ 
    
\State \textrm{Dimension compression}  
    \Statex \quad $H_2(B, D) \leftarrow \textrm{ReduceMean}(H_1)$
    
\State \textrm{Matrix transpose}  
    \Statex \quad $F(D,B) \leftarrow H_2^{\top}$
    \Statex \quad \textrm{Obtain feature matrix $F$} $(D$ nodes with
    feature dimension $B)$  
\end{algorithmic}
\end{algorithm}

Among these, the implementation of the ReduceMean operation may vary depending on the data format being processed, and the following approaches are possible but not exclusive:
\begin{itemize}
\item For $BDHW$-shaped data (batch × depth × height × width), compute the mean value along the height and width dimensions.
\item For $BND$-shaped data (batch × node × dimension), compute the mean along the node dimension.
\end{itemize}
$G_M$ can be viewed as a complex network comprising $D$ nodes, where each node possesses a $B$-dimensional feature vector. The connections between nodes are determined by computing the Euclidean distance between their feature vectors.

\begin{equation} \label{eq:adjacency} 
A_{ij} = 
\begin{cases}
1, & \text{if } \|\mathit{F}_i - \mathit{F}_j\|_2 < \epsilon \\
0, & \text{otherwise}
\end{cases}
\end{equation}

where $A$ is the adjacency matrix of $G_M$, and $\epsilon$ denotes the preset threshold.

\subsection{Quantitative Calculation of Self-Similarity}

After obtaining the feature network, the next step is to quantitatively describe the degree of self-similarity for $G_M$. Drawing on the idea of the box-cover method\cite{kovacs2021comparative}, this paper proposes a modified box-cover algorithm(Appendix A), as shown in Algorithm~\ref{alg:similarity_metric}. The aim is to decompose the feature network $G_M$ into \( N_{\theta} \) local structures (boxes), where nodes within each box are mutually reachable within a limited distance (\( \theta \geq 0 \)).

\begin{algorithm}[htb]
\caption{Self-Similarity Metric $N_{\theta}$ Calculation for Graph Structure $G_M$}
\label{alg:similarity_metric}
\begin{algorithmic}[1]
\Require Graph structure $G_M$ and its feature map $F \in \mathbb{R}^{D \times B}$
\Ensure Metric $N_{\theta}$

\State Compute distance matrix $C \in \mathbb{R}^{D \times D}$: 
\Statex \quad $C_{ij} = \frac{\| F_i - F_j \|_2}{\sqrt{B}}$

\State Define connectivity probability distribution function:
\Statex \quad $p(C \leqslant \theta) = \frac{\sum_{i=1}^{D} \sum_{j=1}^{D} \mathbb{I}(C_{i,j} \leqslant \theta)}{D(D-1)}$ 
\Statex \quad (proportion of connections where $C_{ij} \leqslant \theta$)

\State Calculate metric $N_{\theta}$:
\Statex \quad $N_{\theta} = 1 + (D-1) \log_{D} \left( D + (1-D) p(C \leqslant \theta) \right)$

\State Ensure differentiability through gradient flow:
\Statex \quad $\frac{\partial p(C \leqslant \theta)}{\partial C_{ij}} = \text{fac} \cdot \frac{\partial}{\partial C_{ij}} \text{sigmoid}\left(k(\theta - C_{ij})\right)$
\Statex \quad where $fac$ is the gradient scaling factor, $k$ is the smoothing factor
\end{algorithmic}
\end{algorithm}

When $G_M$ is a standard fractal network, the number of boxes \( N_{\theta} \) in its network decomposition varies with \( \theta \) as follows: 
\begin{equation}
    N_{\theta} \sim (1 + \theta)^{-d_B}
\end{equation}

which means that in logarithmic coordinates, \( \log(N_{\theta}) \) and \( \log(1 + \theta) \) have a linear relationship. Therefore, the degree of self-similarity of $G_M$ can be measured by calculating the deviation of the actual distribution curve from the linear distribution curve.

This paper proposes using the self-similar rate (Self-similar rate, SS\_rate) as an indicator to measure the self-similarity of $G_M$. The formula for calculating SS\_rate is shown in Formula(3-5).
\begin{equation}
\begin{split}
\mathrm{SS\_rate}(G_M) = \frac{2}{D\big(lo(TV) - lo(TZ)\big)} \\ 
\times \int_{TZ}^{TV} \Big| \log N_{\theta} - pf(\theta) \Big| d\log(1 + \theta)
\end{split}
\label{eq:ssrate_split}
\end{equation}

\begin{align}
pf(\theta) &= \frac{lo(TV) - lo(\theta)}{lo(TV) - lo(TZ)} \cdot \log D \label{eq:pf} \\
lo(x) &= \log(1 + x) \label{eq:lo}
\end{align}

Among them, \( D \) represents the number of nodes in $G_M$, while $TZ$ and $TV$ refer to the minimum and maximum values of \( \theta \), respectively. The value of SS\_rate ranges between 0 and 1; the closer it is to 0, the stronger the power-law relationship of \( N_{\theta} \), indicating a stronger self-similarity of $G_M$. Conversely, the closer it is to 1, the weaker the power-law relationship of \( N_{\theta} \), indicating a weaker self-similarity of $G_M$.

\subsection{Loss Function}

Since model M has multiple hidden layers, there will also be multiple ($G_M$). Each $G_M$ possesses its own SS\_rate metric. Constructing a monolithic $G_M$ by combining all hidden layers would incur prohibitive computational complexity. To ensure all $G_M$ participate in training while maintaining efficiency, we randomly select one $G_M$ to calculate the SS\_rate during each iteration of the model. The loss function is accordingly defined as follows:

\begin{equation} \label{eq:ss_reg}
\mathcal{L}_{\text{total}} = \underbrace{\mathcal{L}_{\text{task}}}_{\text{original loss}} + \alpha \cdot \Big\| \mathrm{SS{\_}rate}(G_M^{(k)}) - \gamma \Big\|_2^2
\end{equation}

where $G_M^{(k)}$ represents a randomly sampled feature network from multiple $G_M$ instances of model $M$, and $\gamma$ denotes the target value of self-similarity degree – specifically, the inherent self-similarity level of feature networks after standard training without regularization constraints.

\section{Experiments}
To analyze the self-similarity levels of different models and explore the impact of self-similarity constraints on model performance, this paper designs experiments based on the following three aspects of the models in visual classification tasks:(1) Visualizing and analyzing the feature network ($G_M$) of different models;(2) Investigate the changes in SS\_rate of different models before and after training, and use this as a reference to evaluate the impact of loss function constraints based on SS\_rate on model training performance;  (3) Explore the factors that influence model performance after applying SS\_rate constraints.The source code are publicly available: \\
{\url{https://github.com/qcw-china/Self-similarity-Analysis-in-Deep-Neural-Networks}} \\

\subsection{Datasets and Models}

\textbf{CIFAR-10}\cite{krizhevsky2009learning}, created by Alex Krizhevsky et al. in 2009, consists of 10 object categories (e.g., airplane, automobile, bird) with 6,000 32×32-pixel color images per category, totaling 60,000 images. The dataset is partitioned into 50,000 training samples and 10,000 test samples.

\textbf{CIFAR-100} extends CIFAR-10 with 100 fine-grained categories (e.g., "apple", "brown bear"), each containing 600 images (60,000 total). Maintaining identical 50,000/10,000 train-test partitioning, it introduces 20 superclasses (e.g., "mammals", "vehicles") to enable hierarchical classification tasks.

\textbf{Imagenette}\cite{howard2019imagenette}, curated by Fast.ai in 2019, selects 10 easily distinguishable classes from ImageNet (e.g., "golf ball", "truck") comprising 13,000 images. Its distinguishing feature is higher resolution (default 160×160 pixels) providing richer visual granularity.

This paper adopts the module naming conventions from the timm library for all implemented models, as detailed in Table~\ref{tab:model_modules} which enumerates the specific architectural modules corresponding to each model type.

\begin{table}[htb]
\centering
\small
\begin{tabular}{lcc}
\toprule
\textbf{Model} & \textbf{Modules} \\ 
\midrule
ViT (small)\cite{dosovitskiy2020image} & blocks.0--11 \\ 
ViT (base) & blocks.0--11 \\ 
PoolFormer\cite{yu2022metaformer} & stages.0--3 \\ 
PvT\cite{wang2021pyramid} & stages.0--3 \\ 
\cmidrule[0.8pt](l r){1-2}
Resnet34 \cite{he2016deep}& layer1--4 \\ 
Resnet50 & layer1--4 \\ 
Vgg\cite{simonyan2014very} & features.\{1,4,9,14,19\} \\ 
\cmidrule[0.8pt](l r){1-2}
ResMLP\cite{touvron2022resmlp} & blocks.0--11 \\ 
MLP-Mixer\cite{tolstikhin2021mlp} & blocks.0--11 \\ 
\bottomrule
\end{tabular}
\caption{Model Architectures and Corresponding Modules}
\label{tab:model_modules}
\end{table}

\subsection{$\mathbf{G_M}$ Visualisation}

\begin{figure}[t]
\centering
\includegraphics[width=0.5\textwidth]{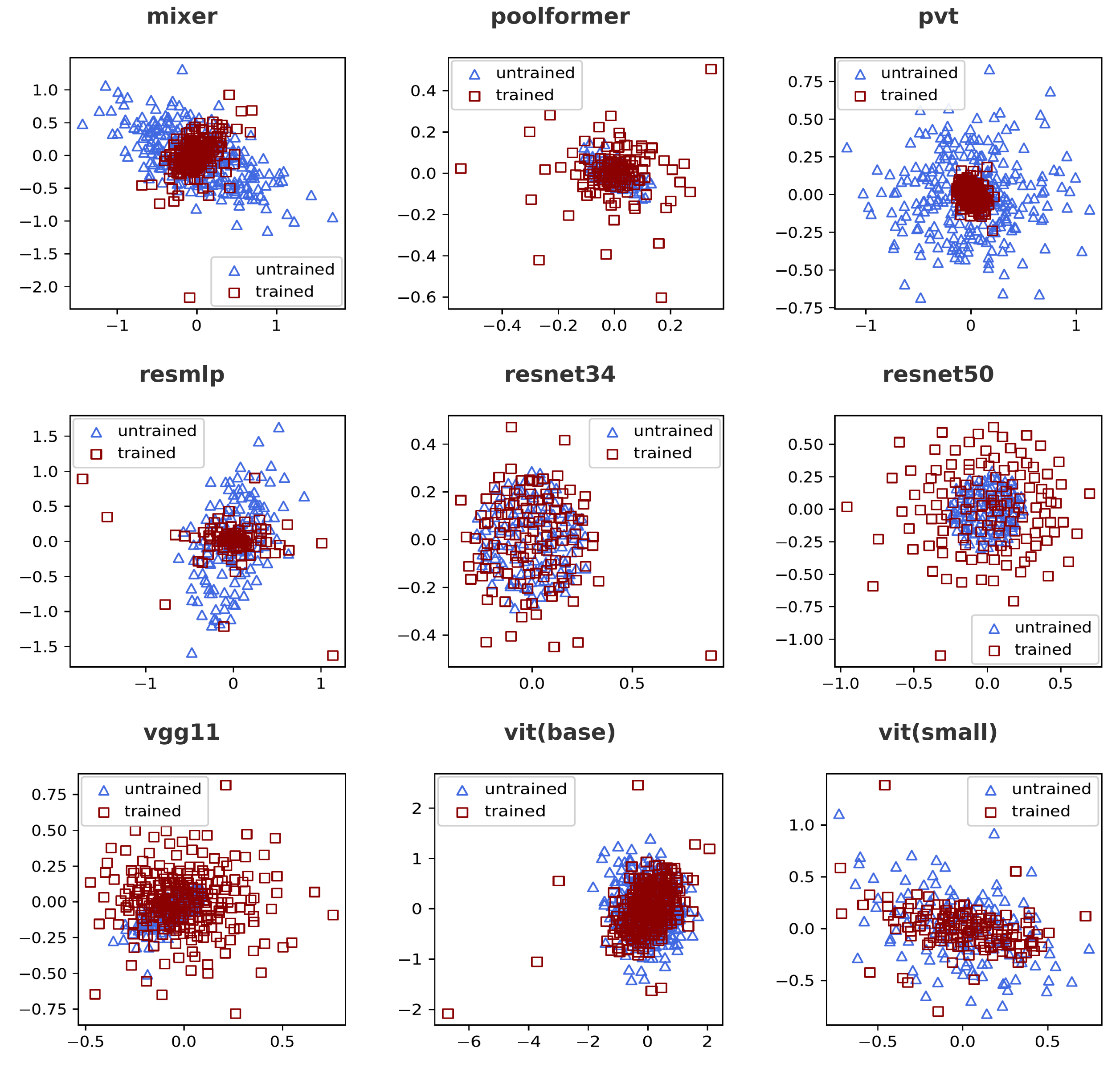} 
\caption{Comparative Visualization of $G_M$ Distributions Across Multiple Architectures Pre- and Post-Training}
\label{fig:output}
\end{figure}

We randomly sample 512 images from the ImageNet training set as model inputs to compute the $G_M$. Since the shallow features close to the input are highly sensitive to pre-training, their differences are significant and follow clear patterns. In contrast, the deep features near the output undergo relatively smaller changes after fine-tuning on the target task using the pre-trained model, making them less sensitive to the pre-training state. Therefore, Figure~\ref{fig:output} presents the visualization results of $G_M$ from the shallow feature networks of different models.The visualization method used in this paper is Multidimensional Scaling (MDS)~\cite{hout2013multidimensional}, a multivariate statistical method that maps objects into a spatial representation based on their similarities or dissimilarities.

From Figure~\ref{fig:output}, it can be observed that the MLP architectures, such as Mixer and ResMLP, as well as the attention architecture ViT, exhibit strong low-dimensional manifold structures in their feature network $G_M$ before and after training, due to their global information aggregation mechanisms. This is specifically reflected in the significant clustering and directionality of the features. 

In contrast, convolutional architectures such as ResNet and Vgg, along with PVT, which have local receptive fields limiting information interaction, show feature network $G_M$ that are more uniformly distributed. This is manifested as isotropic diffusion of features with no dominant direction. The $G_M$ of the PoolFormer model exhibit a strong low-dimensional manifold structure only when the model is untrained, but show a more uniform distribution after pre-training. This is because PoolFormer employs local static mixing operations combined with nonlinear transformations, which disrupt the model's initial global operation mechanism. It can be seen that their core difference lies in the information processing mechanism: architectures with local inductive bias lead to more uniform $G_M$ distributions, whereas architectures with global information aggregation lead to more linear distribution patterns in $G_M$.

\subsection{Quantitative Analysis of Self-Similarity in $\mathbf{G_M}$}
This section delves into the self-similarity of $G_M$. To this end, this study proposes a quantitative metric for self-similarity—the Self-Similarity Rate (SS\_rate). Table~\ref{tab:ssrate_avg} presents the average SS\_rate values of $G_M$ across all hidden layers for each model in both the untrained and pre-trained states.

\begin{table}[htb]
\centering
\begin{tabular}{@{}lcc@{}}
\toprule
\textbf{Model} & \textbf{Pre-train Avg} & \textbf{Post-train Avg} \\
\midrule
\textbf{ViT (small)} & 0.13 & \textbf{0.66} \\
\textbf{ViT (base)} & 0.13 & \textbf{0.46} \\
\textbf{PoolFormer} & 0.34 & \textbf{0.43} \\
\textbf{PvT} & 0.17 & \textbf{0.48} \\
\cmidrule[0.8pt](l r){1-3}
\textbf{Resnet34} & 0.22 & 0.17 \\
\textbf{Resnet50} & 0.22 & \textbf{0.33} \\
\textbf{Vgg} & 0.19 & 0.18 \\
\cmidrule[0.8pt](l r){1-3}
\textbf{ResMLP} & 0.09 & \textbf{0.40} \\
\textbf{MLP-Mixer} & 0.13 & \textbf{0.66} \\
\bottomrule
\end{tabular}
\caption{Model-wise SS\_rate Averages before and after Training}
\label{tab:ssrate_avg}
\end{table}

Table~\ref{tab:ssrate_avg} shows the changes in SS\_rate for all models before and after training. In the majority of models, the SS\_rate increases after training. This indicates that as training progresses, the models gradually transition from a state of high self-similarity to one of lower self-similarity, which aligns with the core optimization process of neural networks evolving from an initial random state toward efficient feature representations. 

Since SS\_rate encodes the relationships between neuron outputs, the high self-similarity observed in the early stages of training reflects a convergence in neurons' activation patterns toward input patterns such as edges and textures. This results in high redundancy in the hidden feature maps across spatial or channel dimensions. As training proceeds, the weights of neurons corresponding to important features are significantly enhanced, while redundant neurons are updated to a lesser extent. This breaks the initial symmetry, reduces the redundancy in hidden feature maps, and consequently lowers the degree of self-similarity.

Next, we design experiments to verify whether constraining SS\_rate during training—specifically by promoting its convergence toward post-training target values—positively impacts model accuracy at convergence. In the experiments, we set the value of the hyperparameter $\alpha$ to the order of magnitude of $10^{-4}$, and each set of experiments is conducted ten times to compute the mean and standard deviation. Our constraint implementation employs a straightforward mean squared error (MSE) loss penalty, leveraging the differentiability and computational efficiency of SS\_rate that we previously emphasized.

\begin{table*}[htb]
\centering
\small 
\scalebox{0.8}{ 
\begin{tabular}{@{}llcc|cc|cc@{}}
\toprule
\multicolumn{2}{c}{\multirow{2}{*}{Model}} & 
\multicolumn{2}{c}{CIFAR10} & 
\multicolumn{2}{c}{CIFAR100} & 
\multicolumn{2}{c}{Imagenette} \\
\cmidrule(lr){3-4} \cmidrule(lr){5-6} \cmidrule(l){7-8}
& & baseline & S\textsuperscript{2}-Con & baseline & S\textsuperscript{2}-Con & baseline & S\textsuperscript{2}-Con \\
\midrule
\multirow{8}{*}{MLP} 
& \multirow{2}{*}{ResMLP$\uparrow$} 
    & 87.35 ($\pm$0.34) & \textbf{88.17 ($\pm$0.09)}
    & 62.16 ($\pm$0.11) & \textbf{64.42 ($\pm$0.16)} 
    & 84.63 ($\pm$0.16) & \textbf{85.89 ($\pm$0.18)} \\
    & & 0.29 ($\pm$0.00) & 0.36 ($\pm$0.00) 
    & 0.24 ($\pm$0.00) & 0.30 ($\pm$0.00) 
    & 0.21 ($\pm$0.01) & 0.37 ($\pm$0.01) \\
    \addlinespace
    
& \multirow{2}{*}{MLP-Mixer$\uparrow$} 
    & 86.81 ($\pm$0.13) & \textbf{88.27 ($\pm$0.13)} 
    & 61.50 ($\pm$0.17) & \textbf{62.53 ($\pm$0.19)} 
    & 82.96 ($\pm$0.14) & 82.16 ($\pm$0.19) \\
    & & 0.15 ($\pm$0.00) & 0.25 ($\pm$0.00) 
    & 0.31 ($\pm$0.00) & 0.29 ($\pm$0.00) 
    & 0.29 ($\pm$0.00) & 0.19 ($\pm$0.00) \\
    \midrule
    
\multirow{12}{*}{CNN} 
& \multirow{2}{*}{Vgg} 
    & 89.98 ($\pm$0.07) & 89.83 ($\pm$0.09) 
    & 60.48 ($\pm$0.18) & 60.06 ($\pm$0.12) 
    & 90.38 ($\pm$0.17) & 90.23 ($\pm$0.16) \\
    & & 0.29 ($\pm$0.01) & 0.33 ($\pm$0.01) 
    & 0.19 ($\pm$0.01) & 0.17 ($\pm$0.01) 
    & 0.30 ($\pm$0.01) & 0.29 ($\pm$0.01) \\
    \addlinespace
    
& \multirow{2}{*}{Resnet34} 
    & 93.90 ($\pm$0.07) & \textbf{94.00} ($\pm$0.10) 
    & 74.23 ($\pm$0.26) & \textbf{74.35} ($\pm$0.31) 
    & 89.20 ($\pm$0.23) & \textbf{89.25} ($\pm$0.14) \\
    & & 0.25 ($\pm$0.01) & 0.24 ($\pm$0.00) 
    & 0.24 ($\pm$0.00) & 0.24 ($\pm$0.00) 
    & 0.19 ($\pm$0.01) & 0.20 ($\pm$0.01) \\
    \addlinespace
    
& \multirow{2}{*}{Resnet50} 
    & 94.11 ($\pm$0.09) & \textbf{94.13} ($\pm$0.13) 
    & 73.77 ($\pm$0.29) & \textbf{73.87} ($\pm$0.22) 
    & 90.50 ($\pm$0.16) & 90.34 ($\pm$0.24) \\
    & & 0.22 ($\pm$0.00) & 0.22 ($\pm$0.00) 
    & 0.24 ($\pm$0.00) & 0.23 ($\pm$0.00) 
    & 0.19 ($\pm$0.01) & 0.17 ($\pm$0.00) \\
    \midrule
    
\multirow{12}{*}{Attention} 
& \multirow{2}{*}{ViT (small)$\uparrow$} 
    & 89.99 ($\pm$0.14) & \textbf{90.50 ($\pm$0.19)} 
    & 65.75 ($\pm$0.16) & \textbf{67.69 ($\pm$0.18)} 
    & 81.71 ($\pm$0.33) & \textbf{83.74 ($\pm$0.28)} \\
    & & 0.49 ($\pm$0.00) & 0.51 ($\pm$0.00) 
    & 0.60 ($\pm$0.00) & 0.52 ($\pm$0.00) 
    & 0.49 ($\pm$0.00) & 0.48 ($\pm$0.01) \\
    \addlinespace
    
& \multirow{2}{*}{ViT (base)$\uparrow$} 
    & 90.29 ($\pm$0.44) & \textbf{93.21 ($\pm$0.23)}
    & 61.40 ($\pm$0.22) & \textbf{67.15 ($\pm$0.22)} 
    & 77.88 ($\pm$0.40) & \textbf{82.93 ($\pm$0.24)} \\
    & & 0.55 ($\pm$0.00) & 0.58 ($\pm$0.00) 
    & 0.41 ($\pm$0.01) & 0.45 ($\pm$0.00) 
    & 0.29 ($\pm$0.00) & 0.39 ($\pm$0.01) \\
    \addlinespace
    
& \multirow{2}{*}{PoolFormer} 
    & 92.85 ($\pm$0.07) & 91.50 ($\pm$0.16) 
    & 71.32 ($\pm$0.17) & 68.56 ($\pm$0.50) 
    & 82.76 ($\pm$0.23) & \textbf{84.70 ($\pm$0.29)} \\
    & & 0.43 ($\pm$0.00) & 0.55 ($\pm$0.00) 
    & 0.46 ($\pm$0.00) & 0.52 ($\pm$0.00) 
    & 0.64 ($\pm$0.00) & 0.48 ($\pm$0.01) \\
    \addlinespace
    
& \multirow{2}{*}{PvT} 
    & 94.55 ($\pm$0.09) & 94.43 ($\pm$0.07) 
    & 74.73 ($\pm$0.09) & 74.45 ($\pm$0.18) 
    & 89.48 ($\pm$0.15) & \textbf{89.64 ($\pm$0.17)} \\
    & & 0.36 ($\pm$0.01) & 0.47 ($\pm$0.01) 
    & 0.44 ($\pm$0.01) & 0.45 ($\pm$0.00) 
    & 0.31 ($\pm$0.01) & 0.35 ($\pm$0.01) \\
\bottomrule
\end{tabular}
} 
\caption{Model performance comparisons across multiple datasets under baseline settings versus SS\_rate-constrained optimization (denoted as S\textsuperscript{2}-Con).}
\label{tab:perf_comparison}
\vspace{-0.5em}
\end{table*}

As demonstrated in Table~\ref{tab:perf_comparison}, the regularization constraint yields significant performance improvements exclusively for ResMLP, MLP-Mixer, and ViT-series models (averaging 1.2-5.3$\%$ gain), while exhibiting negligible impact on convolution-dominated architectures like Vgg/ResNet. This phenomenon contrasts sharply with the universal SS\_rate elevation trend observed in Table~\ref{tab:model_modules}, revealing the model-structure-dependent efficacy of self-similarity constraints.

\subsection{Architectural Self-Similarity: The Decisive Factor in Constraint Efficacy}

Subsequently, we design experiments to validate the correspondence between constrained optimization outcomes and two fundamental properties across different hidden layers: (1) statistical scale invariance \cite{mahoney2019traditional}Table~\ref{tab:stat_scale_invariance} and (2) geometric invariance \cite{grassberger1983measuring}\cite{theiler1990statistical} Table~\ref{tab:geometric_invariance}. Statistical Scale Invariance is quantified by computing feature sparsity and fitting power-law distributions per layer, with invariance strength measured by the standard deviation of power-law exponents across layers. Geometric Invariance is evaluated using the correlation dimension algorithm, where invariance degree is determined by relative fluctuation of fractal dimensions derived from log-log linear fits of correlation integrals.

This investigation aims to establish that intrinsic architectural self-similarity directly governs the effectiveness of our self-similarity constraints. Methodologies for measuring these invariance properties are detailed in the Appendix B,C.

\begin{table}[!htb]
\centering
\small
\begin{tabular}{@{}lcccc@{}}
\toprule
\textbf{Model} & \textbf{CIFAR10} & \textbf{CIFAR100} & \textbf{Imagenette} & \textbf{Avg} \\
\midrule
\textbf{MLP-Mixer} & 0.0758 & 0.0786 & 0.2489 & 0.1344 \\
\textbf{ViT} & 0.1521 & 0.1779 & 0.1722 & 0.1674 \\
\textbf{PoolFormer} & 0.1915 & 0.1886 & 0.3771 & 0.2524 \\
\textbf{Resnet} & 0.2688 & 0.1683 & 0.5158 & 0.3176 \\
\textbf{ResMLP} & 0.3453 & 0.3404 & 0.3404 & 0.3420 \\
\textbf{PvT} & 0.4909 & 0.4506 & 0.7503 & 0.5639 \\
\textbf{Vgg} & 0.7898 & 0.6066 & 0.7482 & 0.7149 \\
\bottomrule
\end{tabular}
\caption{Statistical Scale Invariance Rankings}
\label{tab:stat_scale_invariance}
\end{table}

\begin{table}[!htb]
\centering
\small
\begin{tabular}{@{}lcccc@{}}
\toprule
\textbf{Model} & \textbf{CIFAR10} & \textbf{CIFAR100} & \textbf{Imagenette} & \textbf{Avg} \\
\midrule
\textbf{MLP-Mixer} & 0.2432 & 0.1390 & 0.2270 & 0.2031 \\
\textbf{ResMLP} & 0.2608 & 0.2137 & 0.3417 & 0.2721 \\
\textbf{PoolFormer} & 0.2868 & 0.2126 & 0.3212 & 0.2735 \\
\textbf{ViT} & 0.2456 & 0.2192 & 0.3669 & 0.2772 \\
\textbf{Vgg} & 0.4434 & 0.2422 & 0.2655 & 0.3170 \\
\textbf{Resnet} & 0.3093 & 0.4126 & 0.6399 & 0.4540 \\
\textbf{PvT} & 0.4163 & 0.5249 & 0.6216 & 0.5210 \\
\bottomrule
\end{tabular}
\caption{Geometric Invariance Rankings}
\label{tab:geometric_invariance}
\end{table}

Table~\ref{tab:perf_comparison} shows that the models exhibiting significant performance improvements under the S\textsuperscript{2}-Con regularization constraint—ResMLP, MLP-Mixer, and ViT—are precisely those that demonstrate better statistical scale invariance and geometric invariance (Table~\ref{tab:stat_scale_invariance},~\ref{tab:geometric_invariance}). Statistical scale invariance and geometric invariance are key indicators for evaluating whether a network possesses self-similarity. Therefore, the experimental results indicate that for models that inherently exhibit good self-similarity, imposing such constraints can further enhance their existing performance advantages.

Moreover, the SS\_rate constraint is designed to strengthen self-similarity at the global scale. However, the PoolFormer model, which relies on local features, cannot effectively respond to such global constraints (as analyzed in Figure 2). The Vgg model, which inherently has poor self-similarity, experiences a slight performance drop under the constraint, demonstrating that forcibly altering the local information interaction patterns in models with poor self-similarity can disrupt their optimization process.

\section{Conclusion}

This work proposes $G_M$, a topological feature network constructed from hidden-layer activations – as a new paradigm for complex network representation of neural networks. Transcending conventional structural analysis paradigms,$G_M$ achieves unified modeling across diverse architectures (MLP/CNN/Attention) by capturing dynamic relationships within neuron activation patterns. Experimental results demonstrate that $G_M$ effectively characterizes the dynamics of model parameter distributions. It is observed that the limited information exchange inherent to local receptive fields drives the $G_M$ toward a uniform distribution, while architectures employing global information aggregation exhibit prominent low-dimensional manifold structures. Throughout the training process, the self-similarity of $G_M$ generally diminishes. An self-similarity constraint based on the SS\_rate metric achieved significant performance improvements in models with high self-similarity.

While $G_M$ provides a novel tool for investigating neural network topological properties, the dynamic evolution of self-similarity requires more refined modeling. Current constraint strategies employing fixed self-similarity targets may constrain the autonomous optimization pathways of models. Future research could design time-variant objective functions to dynamically adapt to requirements across distinct training phases, and explore applications of $G_M$ in studying topological characteristics like small-world properties. 

\section{Acknowledgments}

Thank you to the High Performance Computing Center of Xidian University for providing the resource support.

\bibliography{aaai2026}

\appendix
\section{Appendix}
\subsection{A:Derivation from the Standard Box-Covering
Method to Its Variant}
The standard box-covering method, originally developed for analyzing self-similarity in complex networks, faces implementation challenges when applied to $G_M$. Although $G_M$ exhibits a graph structure, its inherent architectural distinctions from conventional complex networks prevent direct application of the box-covering technique.

\textbf{Principle of Box-Cover:}Using the minimum number of boxes to cover the entire network, where nodes within each box satisfy specific conditions (e.g., node-to-node distance does not exceed a given value). By analyzing the power-law relationship between the number of boxes required ($N(\theta)$) at different box sizes (side length l), the fractal dimension $d_B$ is calculated to verify compliance with $N(\theta) \propto l^{-d_B}$

\textbf{The specific process:}

\textbf{Determine Box Size l:}Box size $\theta$ is defined as the maximum shortest-path distance between any two nodes within the same box(When $\theta$ = 3,all node pairs within a single box must satisfy meaning the box can contain subgraphs with diameters up to 2 hops):\begin{equation} \label{eq:box_constraint}
\max_{i,j \in \text{box}} d\left(v_{i}, v_{j}\right) \leq \theta
\end{equation}

\textbf{Calculate N(l) (Minimum Box Count):}\textbf{Burning algorithm:}start from a randomly selected seed node.Then iteratively cover all nodes within distance $\theta-1$.\textbf{Cluster-Growing Algorithm:}expand radially from the seed node as the center and generate boxes through cluster expansion.\textbf{Greedy Algorithm:}select boxes covering the maximum number of uncovered nodes.Then iterate until full network coverage is achieved

\textbf{Compute Fractal Dimension $d_B$:}Calculate the corresponding $ N(\theta) $ for different values of $ \theta $, fit the linear relationship between $ \log N(\theta) $ and $ \log \theta $; the slope is $ -d_B $. If a linear relationship exists, the network exhibits fractal characteristics (self-similarity).

\textbf{Verify Self-Similarity:}Check whether the network statistical properties (such as modularity and node degree distribution) remain consistent at different scales. If the fractal dimension \( d_B \) is constant, then the network exhibits self-similarity.

\( G_M \) only has the feature vectors of each node and does not have the connection relationships between nodes. Therefore, the distance between two nodes cannot be calculated through the paths formed by edges between nodes; it can only be calculated through the distance between feature vectors. Specifically, we use Euclidean distance for calculation, and the corresponding formula in Algorithm 2 is:
\begin{equation}
C_{ij} = \frac{\|F_i - F_j\|_2}{\sqrt{B}}
\end{equation}

In the formula \( N(\theta) \sim \theta^{-d_B} \), if \( \theta \) is less than 1, there may be a risk of numerical overflow. Therefore, we use the corrected formula \( N(\theta) \sim (1 + \theta)^{-d_B} \) to analyze the correlation. This means we are examining the linear relationship between \( \log N(\theta) \) and \( \log (1 + \theta) \).

Furthermore, if we want to constrain the self-similarity of the model using a loss function, it is not possible to use iterative algorithms such as the burning method or clustering methods to calculate the actual \( N(\theta) \) values. Therefore, we adopt an approximate simulation method to compute the simulated \( N(\theta) \) values. The corresponding formula in Algorithm 2 is:
\begin{equation}
N_{\theta} = 1 + (D - 1) \log_D (D + (1 - D) p(C \leq \theta))
\end{equation}

By directly using the connection density under the current threshold \( \theta \) to simulate \( N_{\theta} \), we can quickly run the modified box-cover algorithm. This allows us to subsequently apply it in the calculations of the loss function and SS\_rate.

\subsection{B:Computational Method for Scale Invariance Statistics}
Statistical scale invariance is a key characteristic of self-similarity in complex systems, originating from fractal geometry theory. According to fractal theory, the statistical properties of scale-invariant systems maintain power-law relationships across different observation scales. Based on this principle, we propose the following quantitative method:

\textbf{Feature Matrix Processing}:for the feature matrix \( F^{(l)} \in \mathbb{R}^{(D, B)} \) extracted from the \( l \)-th layer, calculate its covariance matrix:
\begin{equation}
\Sigma^{(l)} = \frac{1}{B-1} (F^{(l)})^T F^{(l)}
\end{equation}
The covariance matrix \( \Sigma^{(l)} \) encodes the linear correlation structure between feature dimensions, and its spectral properties reflect the geometric invariance of the feature space.

\textbf{Spectral Analysis}:calculate the eigenvalue spectrum of the covariance matrix:
\begin{equation}
\lambda^{(l)} = \text{eigvals}(\Sigma^{(l)})
\end{equation}
The eigenvalues \( \lambda^{(l)} \) represent the variance intensity of the feature space along the orthogonal basis directions. The distribution pattern of these eigenvalues reflects the scale characteristics of hierarchical structures.

\textbf{Power-Law Distribution Fitting:}assume that the eigenvalue distribution follows a power-law distribution:
\begin{equation}
    P(\lambda) \propto |\lambda|^{-\gamma} \quad (\lambda > \lambda_{\min})
\end{equation}
Use Maximum Likelihood Estimation (MLE) to solve for the index \( \gamma_l \):
\begin{equation}
    \hat{\gamma}_l = 1 + D \left[ \sum_{k=1}^{D} \ln \frac{\lambda_k}{\lambda_{\min}} \right]^{-1}
\end{equation}

\textbf{Measurement of Scale Invariance:}calculate the standard deviation of all hierarchical power-law indices:
\begin{equation}
    \sigma_{\alpha, \gamma} = \sqrt{\frac{1}{|L|} \sum_{l \in L} (\gamma_l - \bar{\gamma})^2}, \quad \bar{\gamma} = \frac{1}{|L|} \sum_{l \in L} \gamma_l
\end{equation}
Where $L$ denotes the number of model layers.A smaller value of \( \sigma_{\alpha, \gamma} \) indicates stronger statistical scale invariance between hierarchical levels.
\subsection{C:Geometric Invariance Calculation Method}
Geometric invariance analysis aims to quantify the self-similar properties of fractal structures in feature space across different network layers.

\textbf{Feature Matrix Dimensionality Reduction:}to overcome the challenges of analyzing high-dimensional feature spaces, we use the UMAP algorithm for topologically preserving dimensionality reduction of latent features:
\begin{equation}
X^{(i)} = \text{UMAP}\left(\hat{F}^{(i)}, n_{\text{components}}, n_{\text{neighbors}}, \text{min\_dist}\right)    
\end{equation}
Here, \( n_{\text{components}} \) is the target dimension after reduction, \( n_{\text{neighbors}} \) and \( \text{min\_dist} \) are parameters that control the preservation of local and global structures.

\textbf{Calculation of Correlation Dimension:}for the dimensionality-reduced features \( X^{(i)} \), calculate the correlation dimension using the correlation integral:
Calculate pairwise Euclidean distances:
\begin{equation}
  d_{ij} = \| X_i^{(i)} - X_j^{(i)} \|_2  
\end{equation}
Define the correlation integral function:
\begin{equation}
   T(r) = \frac{1}{N(N-1)} \sum_{i=1}^{N} \sum_{\substack{j=1 \\ j \neq i}}^{N} \Theta(r - d_{ij}) 
\end{equation}
where \( \Theta \) is the Heaviside step function.
Fit a power-law relationship in a log-log coordinate system:
\begin{equation}
D_{\text{corr}}^{(i)} = \text{slope}\left(\log T(r)\right)
\end{equation}

\textbf{Fluctuation Analysis of Fractal Dimensions:}calculate the normalized fluctuation of correlation dimensions across hierarchical layers:
\begin{equation}
 \Delta = \frac{\max_{l \in L} D_2^{(l)} - \min_{l \in L} D_2^{(l)}}{\text{mean}_{l \in L} D_2^{(l)}}   
\end{equation}
Here, \( L \) represents the set of network layers. A smaller value of \( \Delta \) indicates that the feature space maintains geometric structural consistency across different abstract hierarchical levels, implying stronger self-similarity.

\end{document}